%% file: main.tex
\documentclass[sigconf,nonacm]{acmart}

\setcitestyle{numbers,square,sort&compress}

\usepackage{amsmath,amssymb,amsfonts}
\usepackage{graphicx}
\graphicspath{{pic/}{figures/}}
\usepackage{booktabs}
\usepackage{multirow}
\usepackage{array}
\usepackage{xcolor}
\usepackage{microtype}
\usepackage{enumitem}
\usepackage{afterpage}

\renewcommand\footnotetextcopyrightpermission[1]{}
\setcopyright{none}

\newcommand{\irepa}{\textsc{iREPA}}

\newcommand{\RB}{\textsc{RetroBridge}}
\newcommand{\synbridge}{\textsc{SynBridge}}
\newcommand{\repa}{\textsc{REPA}}
\newcommand{\reg}{\textsc{REG}}
\newcommand{\USPTOK}{\textsc{USPTO-50k}}
\newcommand{\ours}{\textsc{GRG}}

\newcommand{\nsamples}{100}
\newcommand{\Tsteps}{500}

\title{Representation-Guided Discrete Molecular Graph Retrosynthesis}

\author{Jiahai Huang}
\affiliation{
  \institution{Sun Yat-sen University}
  \city{Guangzhou}
  \country{China}
}

\author{Anjie Qiao}
\affiliation{
  \institution{Sun Yat-sen University}
  \city{Guangzhou}
  \country{China}
}

\author{Zhen Wang}
\affiliation{
  \institution{Sun Yat-sen University}
  \city{Guangzhou}
  \country{China}
}

\author{Defu Lian}
\affiliation{
  \institution{University of Science and Technology of China}
  \city{Hefei}
  \country{China}
}

\author{Yutong Lu}
\affiliation{
  \institution{Sun Yat-sen University}
  \city{Guangzhou}
  \country{China}
}

\begin{abstract}
Stochastic process–based molecular graph generators have become the state of the art for template-free single-step retrosynthesis. However, these models are typically trained only on product–reactant pairs, thereby acquiring chemistry-relevant representations in an indirect and implicit manner. Meanwhile, recent advances in computer vision demonstrate that offering representation guidance to a generator can effectively distill semantics from pretrained encoders into DiTs, substantially improving both convergence and generation quality. Whether similar gains extend to the retrosynthesis task, and what graph-specific design choices can make them work, remains an open question.
To address these questions, we conduct a systematic empirical study over a unified design space spanning teacher molecular representations, endpoint and granularity choices, injection depths in the denoiser, correspondence strategies and guidance scheme. Guided by these considerations, we develop \textbf{G}raph-oriented \textbf{R}epresentation \textbf{G}uidance (\textbf{\ours}), which achieves 58.6 / 77.2 / 83.4 / 87.1 top-1 / 3 / 5 / 10 accuracy on \USPTOK{}, while increasing diversity to 15.5, both substantially outperforming the adopted base generator.
Notably, \ours{} consistently improves all top-$k$ metrics in out-of-distribution settings, suggesting that representation guidance facilitates the acquisition of intrinsic chemical semantics.
Meanwhile, the introduced representation guidance reduces the number of epochs by 35\% and the wall-clock time by 30\% to reach comparable performance.
In addition, we introduce a simple yet effective representation-similarity-based reranking mechanism, which further improves the top of the ranked list without training an additional verifier.
\end{abstract}

\setlist[itemize]{leftmargin=*, noitemsep, topsep=2pt, parsep=0pt, partopsep=0pt}
\setlist[enumerate]{leftmargin=*, noitemsep, topsep=2pt, parsep=0pt, partopsep=0pt}
\begin{document}

\maketitle

\begin{figure*}[t]
\centering
\includegraphics[width=\textwidth]{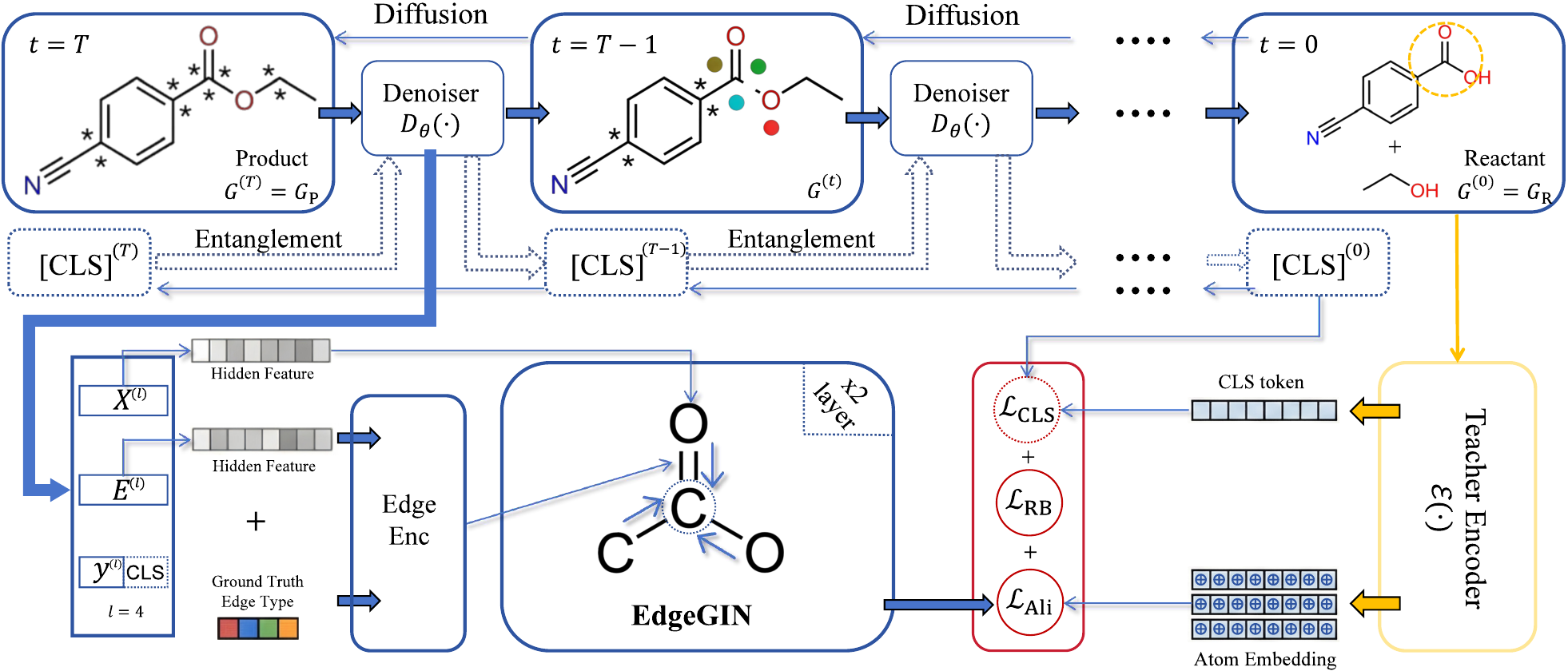}
\caption{Overview of representation guidance for stochastic-process retrosynthesis and our node--edge projector}
\Description{Overview of representation guidance for stochastic-process molecular-graph retrosynthesis and the two guidance schemes (alignment and entanglement), including the node--edge coupled EdgeGIN projector used in GRG.}
\label{fig:main}
\end{figure*}

\input{sections/intro}
\input{sections/related}
\input{sections/method}

\begin{figure*}[t]
  \centering
  \includegraphics[width=\textwidth]{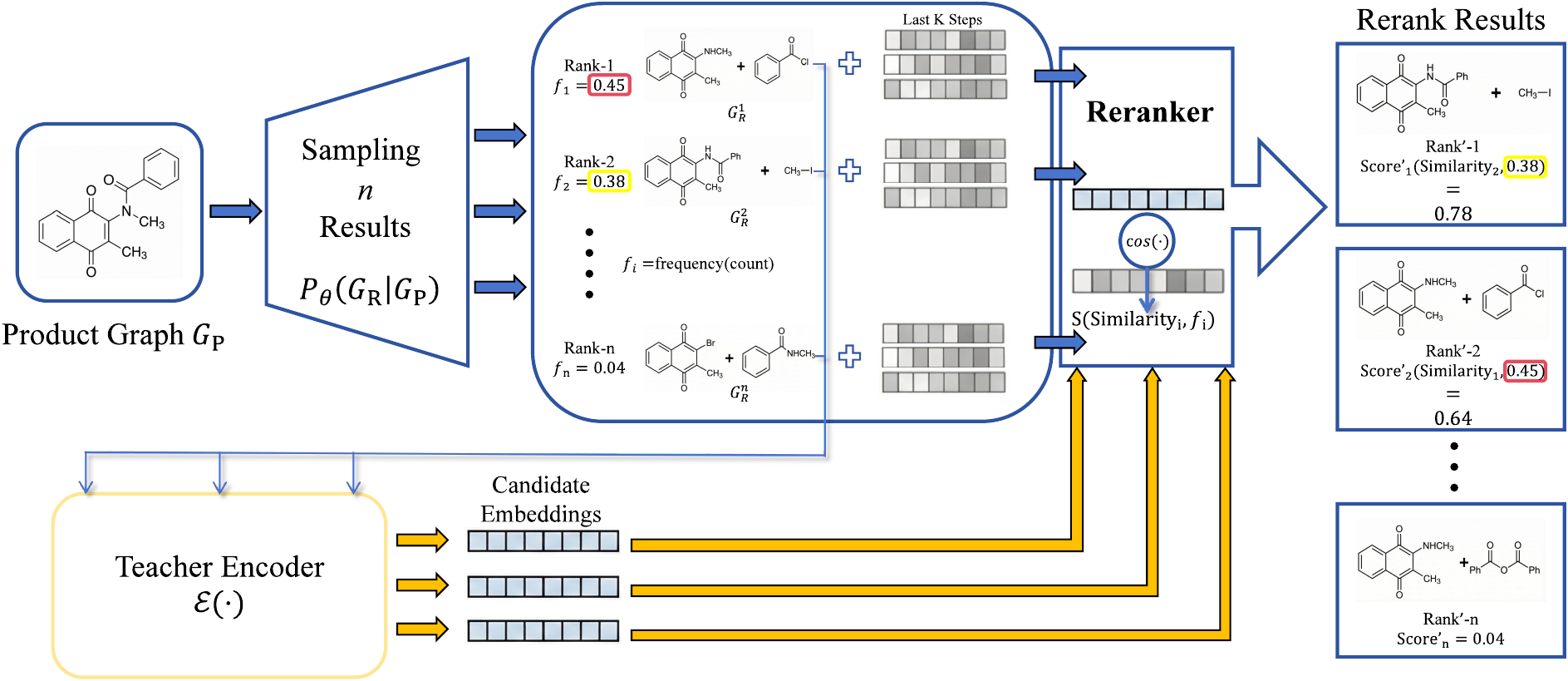}  
  \caption{Reranking sampled candidates by frequency and representation similarity.}
  \label{fig:rerank}
\end{figure*}

\input{sections/tts}
\input{sections/results}

\input{sections/limitations}

\input{sections/conclusion}

\bibliographystyle{ACM-Reference-Format}
\bibliography{references}

\appendix
\input{sections/appendix}

\end{document}

%% file: sections/intro.tex
\section{Introduction}
\label{sec:intro}
Retrosynthesis prediction---decomposing a given product into plausible precursor sets---is a central primitive in synthesis planning and drug discovery~\cite{segler2018planning,striethkalthoff2020machinelearning}.
Modern retrosynthesis systems span a spectrum of reliance on prior chemical knowledge: template-based methods select from a predefined reaction template set~\cite{dai2019gln,chen2021localretro}; semi-template methods first identify reaction centers or synthons and then complete reactants~\cite{yan2020retroxpert,sacha2021megan,wang2021retroprime}; and template-free methods directly generate reactants end-to-end, offering the best scalability to expanding reaction corpora~\cite{zheng2019scrop,schwaller2019moleculartransformer,wan2022retroformer}.

Within template-free modeling, most work represents molecules as SMILES strings~\cite{weininger1988smiles} and frames retrosynthesis as translation~\cite{schwaller2019moleculartransformer,tetko2020augmented,kim2021tied,wan2022retroformer}.
Graph-based approaches instead operate on molecular graphs and can preserve structural constraints more directly~\cite{shi2020g2g,somnath2021graphretro}.
Recently, \emph{stochastic process--based generative models} have become the state of the art for graph-to-graph retrosynthesis: rather than predicting the full reactant set in one shot, they generate reactants through a sequence of stochastic refinement steps instantiated as discrete diffusion, Markov bridges, or discrete-flow formulations~\cite{austin2021d3pm,vignac2023digress,igashov2024retrobridge,lin2025synbridge}.
In this work, we focus on improving such models.

Despite their promising performance, stochastic process--based retrosynthesis models learn the conditional transport induced by paired product–reactant supervision.
However, such supervision only \emph{implicitly} shapes intermediate representations of the model, which may fall short of encoding chemistry-relevant abstractions (e.g., functional groups, local atomic environments, or reactivity patterns) that remain informative under noise and help determine which refinements are plausible.
Meanwhile, recent work in vision has shown that distilling representations from pretrained encoders into diffusion/flow Transformers---via representation alignment (\repa{})~\cite{yu2024repa} and representation entanglement (\reg{})~\cite{wu2025reg}---can close this semantic gap and enhance both convergence speed and generation quality.
This motivates a natural question: can representation guidance also improve \emph{discrete molecular-graph retrosynthesis}, and what graph-specific design choices are needed for it to work reliably?

Adapting guidance from continuous vision models to discrete molecular graphs raises several \emph{design questions}: teacher choice, endpoint/granularity decisions, how to incorporate bond-aware structure beyond isolated atom tokens, and robustness of the guidance scheme under \emph{categorical corruption}---the forward noising that corrupts \emph{categorical} variables (here, atom/bond types) via discrete transition matrices rather than adding Gaussian noise.
We formalize these questions as study axes in Sec.~\ref{sec:study_axes}.

\noindent\textbf{Our work.}
We conduct a systematic empirical study on \USPTOK{} using a representative Markov-bridge retrosynthesis generator as the primary backbone~\cite{igashov2024retrobridge}, and additionally validate key conclusions on a discrete-flow model~\cite{lin2025synbridge}.
From an overarching perspective, we study two guidance schemes:
(i) \emph{representation alignment} (\repa{}), which aligns intermediate node-/graph-level hidden states with teacher embeddings, and
(ii) \emph{representation entanglement} (\reg{}), which jointly corrupts and denoises an auxiliary semantic token together with the graph state to couple semantics and generation.
To make these schemes work on molecular graphs under categorical corruption---and to balance accuracy and diversity in one-to-many reaction spaces---we further explore a graph-oriented design space and systematically examine combinations of:
teacher encoders/representations (ECFP/Morgan fingerprints, RMAT, Uni-RXN, Uni-Mol+),
guidance location (reactant-only vs.\ dual-endpoint),
alignment granularity (graph-level vs.\ node-level),
injection depth (which denoiser layer to align),
guidance scheme (\repa{}/\reg{}),
and correspondence strategies that handle node--bond-coupled guidance.
Guided by these graph-specific considerations, we develop \textbf{G}raph-oriented \textbf{R}epresentation \textbf{G}uidance (\textbf{\ours}), a structure-aware alignment design for stochastic-process retrosynthesis generators.

Beyond training, we show that representation similarity provides a lightweight inference-time verification signal: without training a separate verifier, we rerank sampled candidates by representation similarity.

\noindent\textbf{Contributions.} We summarize our main contributions as follows:
\begin{itemize}[leftmargin=*]
  \item We propose \ours, a graph-oriented representation guidance framework for discrete molecular graph retrosynthesis. It adapts representation guidance to stochastic process--based graph generators and introduces graph-specific designs (e.g. correspondence handling and node--edge--aware intermediate alignment) that are compatible with categorical corruption.
  \item We provide a systematic design study that compares teacher representations, guidance location, alignment granularity, injection depth, guidance schemes, and correspondence strategies, yielding practical guidance on how to effectively apply representation guidance for retrosynthesis under discrete corruption.
  \item We show that representation similarity can serve as a lightweight inference-time verification signal: a simple representation-similarity reranking improves the head of the ranked list with minimal extra logic and without training an additional verifier. Our best configuration achieves state-of-the-art performance.

\end{itemize}

%% file: sections/related.tex
\section{Background and Related Work}
\label{sec:related}

\noindent\textbf{Single-step retrosynthesis: template-based, semi-template, and template-free.}
Template-based methods rely on reaction templates and cast retrosynthesis as classification or retrieval over templates~\cite{dai2019gln,chen2021localretro}.
Semi-template pipelines decompose the task into intermediate subproblems such as reaction-center identification, synthon generation, and synthon-to-reactant completion~\cite{yan2020retroxpert,sacha2021megan,wang2021retroprime}.
Template-free methods generate reactants end-to-end and are scalable to large reaction datasets and diverse chemical spaces~\cite{segler2018planning,striethkalthoff2020machinelearning}.
Most template-free methods operate on SMILES~\cite{weininger1988smiles} and cast retrosynthesis as SMILES-to-SMILES translation using sequence-to-sequence models (often Transformer-based)~\cite{schwaller2019moleculartransformer,zheng2019scrop,tetko2020augmented,kim2021tied,wan2022retroformer,han2024iterative}.
Graph-based methods instead aim to preserve molecular structure and typically treat retrosynthesis as conditional graph generation~\cite{shi2020g2g,somnath2021graphretro,tu2022graph2smiles}.
Our work focuses on \emph{stochastic-process graph generators} and studies how representation guidance should be designed for discrete molecular-graph retrosynthesis.

\noindent\textbf{Discrete diffusion, bridges, and flows for graph generation.}
Generative diffusion models~\cite{sohl2015diffusion,ho2020ddpm} have been extended to discrete state spaces~\cite{austin2021d3pm} and to graphs~\cite{vignac2023digress}.
For retrosynthesis, recent work has explored diffusion-like iterative generation on molecular graphs, achieving state-of-the-art accuracy and validity~\cite{wang2023retrodiff,laabid2024diffalign}.
\RB{} formulates retrosynthesis as sampling from a Markov bridge distribution that explicitly couples the product and reactant endpoints~\cite{igashov2024retrobridge}.
In addition to diffusion/bridges, discrete flow formalisms are also emerging for bidirectional retrosynthesis and reaction-state modeling~\cite{lin2025synbridge}.
Our empirical study injects representation guidance into both \RB{} and \synbridge{}.

\noindent\textbf{Representation guidance: alignment and entanglement.}
Diffusion/flow Transformers~\cite{peebles2023dit} expose intermediate hidden states along the denoising trajectory, enabling auxiliary objectives that improve convergence speed and generation quality. \emph{Representation alignment} (\repa{}) aligns intermediate denoising features with embeddings produced by a pretrained encoder~\cite{yu2024repa}. \emph{Representation entanglement} (\reg{}) models a joint distribution over the generated sample (e.g., an image; in our setting, the reactant graph) and an auxiliary token, jointly corrupting/denoising the token and the sample to couple semantics and generation~\cite{wu2025reg}.
Meanwhile, a key insight from recent studies is that alignment effectiveness may be driven not only by global semantic quality, but also by how well \emph{local structure} is preserved in the representation space. For example, \irepa{}~\cite{singh2025irepa} shows that enhancing the transfer of spatial structure (via a more structure-preserving projection layer and normalization on encoder features) can significantly improve convergence in diffusion Transformer training. This perspective motivates our graph-specific design: in molecular graphs, chemically meaningful structure is carried by \emph{bonded neighborhoods} and fragment-level patterns, suggesting that guidance schemes should explicitly account for node--edge coupling rather than aligning atom tokens in isolation.

\noindent\textbf{External molecular encoders and representations.}
In this work, we consider a spectrum of molecular representations used as teachers or entangled tokens.
Morgan/ECFP fingerprints~\cite{rogers2010ecfp} are a strong hand-engineered baseline.
Learned encoders include RMAT~\cite{maziarka2021rmat} and Uni-Mol+~\cite{zhou2023unimol}, which provide transferable molecule embeddings.
We also use Uni-RXN~\cite{qiang2023unirxn}, a reaction-aware representation model bridging reaction pretraining and conditional generation, as our primary ``teacher'' encoder.
Because these teachers differ in modality (2D vs.\ 3D), supervision granularity (graph vs.\ atom), and inductive bias, we treat teacher choice as a first-class axis rather than assuming a universally optimal encoder.

\noindent\textbf{Evaluation, validity checks, and inference-time reranking.}
Standard single-step retrosynthesis evaluation reports top-$k$ exact-match accuracy on benchmarks such as \USPTOK{}~\cite{dai2019gln,wan2022retroformer}.
Forward-model round-trip checks using the Molecular Transformer~\cite{schwaller2019moleculartransformer,schwaller2020pathways} are widely used to assess plausibility.
For generative models that produce many samples per product, ranking heuristics such as frequency among samples are commonly used~\cite{igashov2024retrobridge}.
We show that representation similarity additionally enable a lightweight inference-time reranking signal without training a separate verifier.

%% file: sections/method.tex
\section{Empirical Study Design}
\label{sec:method}

This section outlines the experimental protocol, covering the task formalism (Sec.~\ref{sec:task}) and design questions for representation guidance (Sec.~\ref{sec:study_axes}).

\subsection{Task setup and evaluation protocol}
\label{sec:task}
\label{sec:exp_setup}
We focus on template-free, single-step retrosynthesis mapping a product graph $G_P$ to reactants $G_R$. Following prior work~\cite{igashov2024retrobridge,vignac2023digress}, reactants are treated as a single disconnected graph padded to $N$ nodes.

We represent a molecular graph as $G=(X,E)$, where $X \in \mathbb{R}^{N \times K_a}$ is the node (atom) feature matrix and $E \in \mathbb{R}^{N \times N \times K_b}$ is the edge (bond) feature tensor.
Here $K_a$ and $K_b$ denote the dimensionalities of the (discrete) atom- and bond-feature vocabularies (e.g., one-hot atom types and one-hot bond types), respectively.
Given paired training examples $(G_P, G_R)\sim p_{\text{data}}$, we train a parameterized conditional model $p_\theta(G_R \mid G_P)$ to approximate the unknown conditional distribution $p_{\text{data}}(G_R \mid G_P)$.

\paragraph{Datasets.}
We utilize the standard \USPTOK{} benchmark~\cite{schneider2016uspto} with established splits~\cite{dai2019gln,somnath2021graphretro} and further evaluate OOD generalization on \textsc{USPTO-50k-cluster}~\cite{qiao2025rarb} (Appendix~\ref{sec:appendix_cluster}).

\paragraph{Sampling and metrics.}
For each product, we sample $n=\nsamples$ unique candidates from $p_\theta$, ranked by frequency.
We report top-$k$ accuracy, round-trip metrics via the Molecular Transformer~\cite{schwaller2019moleculartransformer,schwaller2020pathways}, and diversity.
See Appendix~\ref{sec:metrics} for definitions.

\paragraph{Backbones and protocol.}
RetroBridge is a state-of-the-art stochastic-process molecular-graph retrosynthesis model; accordingly, we adopt \RB{}~\cite{igashov2024retrobridge} as our primary generator to investigate representation guidance, with additional validation on \synbridge{}~\cite{lin2025synbridge}. Following official implementations and consistent preprocessing, we provide details in Appendix~\ref{sec:impl}. Code is available at \url{https://anonymous.4open.science/r/rgdmg7f3c9a/}.

\subsection{Key design questions and study axes}
\label{sec:study_axes}
Adapting representation guidance from continuous generative models to \emph{discrete} molecular-graph retrosynthesis is not a drop-in change. Retrosynthesis generators operate on \emph{categorical} atom/bond variables~\cite{austin2021d3pm,vignac2023digress} and the task is inherently \emph{one-to-many}.
This creates several task-specific challenges.
We therefore undertake an empirical design for representation guidance in discrete molecular-graph retrosynthesis, and organize our study around the following interrelated questions:

\smallskip
\noindent (Q1) \textbf{Teacher representations: which molecular representations are effective teachers?} \\
Unlike vision, retrosynthesis lacks a universally optimal molecular encoder:
existing encoders differ in modality (2D vs.\ 3D), pretraining objective, and representational granularity, and their suitability as teachers remains uncertain.
This calls for a systematic comparison of teacher representations rather than assuming a single best teacher. \\
\emph{Design axis:} choice of pretrained encoder $\mathcal{E}(\cdot)$. \\
\emph{Choices:} Morgan/ECFP fingerprints~\cite{rogers2010ecfp}, RMAT~\cite{maziarka2021rmat}, Uni-Mol+~\cite{zhou2023unimol}, and Uni-RXN~\cite{qiang2023unirxn}. 

\smallskip
\noindent (Q2) \textbf{Where and at what granularity should guidance be applied in one-to-many reaction spaces?} \\
In practice, many molecule encoders naturally expose graph-level embeddings (fingerprints, \texttt{[CLS]}/pooling), while atom-level embeddings, when available, are often intermediate and not always a clean ``atom token $\rightarrow$ atom embedding'' interface.
This makes graph-level guidance broadly applicable, but it can be overly coarse: a single global vector does not specify \emph{where} to edit the reactant graph, potentially under-constraining local chemistry while over-regularizing the overall reactant distribution.
At the graph level, a natural starting point is \emph{dual-endpoint} alignment: teacher embeddings can be computed for both product and reactants, and many stochastic-process formulations (especially bridge-style views) make it tempting to constrain both endpoints for semantic consistency. With teacher representations available, the choice of alignment endpoint and injection layer remains non-trivial.
Adding global constraints to the already-observed product side may over-regularize conditional generation, narrowing the candidate distribution and hurting diversity (or large-$k$ performance) in one-to-many reaction spaces.
Moreover, even when atom-level targets are available, moving from graph-level to node-level guidance is not obviously free: finer constraints may tilt the accuracy--diversity trade-off, for example by biasing the model toward a single decomposition and reducing diversity or large-$k$ coverage on multi-reaction-center reactions.
Whether such trade-offs occur in discrete retrosynthesis must be tested empirically.
In Sec.~\ref{sec:results_main}, we show that node-level guidance not only avoids the feared degradation but also improves both accuracy and diversity. \\
\emph{Design axes:} (i) endpoint choice, (ii) granularity of the guided representation, and (iii) injection depth in the denoiser. \\
\emph{Choices:} guiding both endpoints (product+reactant) vs.\ guiding the generated side only (reactant-only);
graph-level guidance on the global token $y$ vs.\ node-level guidance on atom tokens $X$;
and injection depth by aligning different denoiser layers (Sec.~\ref{sec:layer_results}).

\smallskip
\noindent (Q3) \textbf{Graph-specific structure: do we need designs beyond isolated atom tokens?} \\
Unlike images with patch tokens, molecular graphs have no canonical local unit.
Even with node-level guidance, a single atom embedding can be too local to represent fragment-level semantics (e.g., functional groups or ring systems), and chemical meaning depends critically on bonds and valence constraints.
This suggests that effective guidance should incorporate structure-aware intermediate units that couple nodes and bonds. \\
\emph{Design axis:} whether and how to incorporate bond-aware local topology into representation transfer. \\
\emph{Choices:} a plain per-node MLP projector (atom-token alignment) vs.\ structure-aware node--edge coupling (our \ours{}),
where the projector aggregates bonded neighborhoods via message passing~\cite{xu2019gin} and enhances relative information among atoms with per-molecule instance normalization~\cite{ulyanov2016instance}.

\smallskip
\noindent (Q4) \textbf{Guidance scheme: which scheme is robust under categorical corruption?} \\
Guidance schemes may behave differently under categorical corruption.
Most representation guidance schemes are developed for continuous diffusion/flow models~\cite{yu2024repa,wu2025reg}, where small guidance-induced perturbations typically change samples smoothly and can be corrected over later denoising steps.
In discrete graph generation, however, the denoiser predicts categorical distributions over atom and bond types~\cite{austin2021d3pm,vignac2023digress}.
Because guidance affects logits through intermediate features, even a small logit shift can flip the most probable category (e.g., atom/bond type), potentially violating chemical constraints and pushing the trajectory into hard-to-recover regions.
Thus, it is an empirical question which guidance scheme is robust without harming validity or collapsing diversity. \\
\emph{Design axis:} how semantics is injected into the generator under categorical corruption. \\
\emph{Choices:} representation alignment (\repa{})~\cite{yu2024repa} vs.\ representation entanglement (\reg{})~\cite{wu2025reg}.
We compare these schemes through accuracy--diversity trade-offs.

\section{Representation Guidance Design}
\label{sec:guidance_design}
Guided by the study axes in Sec.~\ref{sec:study_axes}, we now detail how representation guidance is
instantiated within stochastic-process molecular-graph generators. We start from a unified denoiser
interface and the intermediate states exposed along the refinement trajectory
(Sec.~\ref{sec:guidance}). We then discuss representation alignment (\repa{}) and entanglement (\reg{}), and introduce \ours{}. This graph-oriented variant couples node and bond context to ensure stable transfer of chemical semantics.

\subsection{Stochastic-process graph generators}
\label{sec:guidance}
All guidance instantiations depend on a stochastic-process graph generator that iteratively refines a graph state over steps $t=1,\dots,T$ and exposes intermediate hidden states.
We denote its denoiser as $D_\theta(\cdot)$, implemented as a Graph Transformer.
At Transformer layer $\ell$, the denoiser maintains and can expose intermediate representations:
\emph{node representations} $X^{(\ell)}_t \in \mathbb{R}^{N\times d_x}$,
\emph{edge representations} $E^{(\ell)}_t \in \mathbb{R}^{N\times N\times d_e}$,
and a \emph{global token} $y^{(\ell)}_t \in \mathbb{R}^{d_y}$.
Representation guidance attaches lightweight heads on top of these intermediate states and adds auxiliary objectives.

\subsection{Representation alignment (\repa{}) for retrosynthesis}
\label{sec:repa_method}
We adapt representation alignment to discrete graph generation by attaching auxiliary heads to intermediate denoiser features and aligning them to embeddings produced by a pretrained teacher encoder $\mathcal{E}(\cdot)$.

\subsubsection{Graph-level alignment on the global token $y$}
Some teachers naturally output only a single graph embedding (e.g., ECFP), which motivates graph-level alignment as the first instantiation.

Let $\mathcal{E}_{\texttt{CLS}}(M)$ denote the teacher's graph-level embedding of molecule $M$ (either the encoder's dedicated \texttt{[CLS]} output or a single-vector representation such as ECFP).
For a reactant set $G_R$ with multiple components $\{M_1,\dots,M_{K_R}\}$, we define a set embedding by summing per-molecule vectors:
\begin{equation}
    h_R = \sum_{j=1}^{K_R} \mathcal{E}_{\texttt{CLS}}(M_j),
    \label{eq:set_embed}
\end{equation}
and obtain $h_P$ for the product analogously.
We apply standard normalization (Appendix~\ref{sec:impl}) and attach an MLP projection head $g_{\phi_\star}$ to $y^{(\ell_\star)}_t$:
$\hat{h}_{\star,t} = \text{Norm}(g_{\phi_\star}(y^{(\ell_\star)}_t))$.
The alignment loss is cosine distance: $\mathcal{L}_{\text{align-}y} = \sum_{\star \in \mathcal{S}} (1-\cos(\hat{h}_{\star,t}, \tilde{h}_\star))$, where $\mathcal{S} \subseteq \{P, R\}$ controls endpoint choice.

\subsubsection{Node-level alignment on $X$: atom-level targets}
Graph-level alignment can be overly coarse.
We therefore study node-level alignment that supervises individual atom representations.
We require atom-aligned teacher features; in our study Uni-RXN provides per-atom embeddings in a canonical RDKit atom order.
During preprocessing we cache the node-order mapping and \emph{reorder} teacher atom embeddings to match the student node indices; we therefore denote the aligned teacher target for student node $i$ directly as $\tilde{h}_{i}$ (details in Appendix~\ref{sec:appendix_order}).
Let $\mathcal{V}_R$ denote the set of valid reactant atom indices (excluding padded/dummy slots).

We attach a node-wise MLP $g_{\phi_X}$ to the student's node features $X^{(\ell)}_t$.
Let $\hat{h}_{t,i} = \text{Norm}(g_{\phi_X}(X^{(\ell)}_{t,i}))$ and $\tilde{h}_{i}$ denote the normalized teacher target.
We apply cosine-distance alignment over valid reactant atoms: $\mathcal{L}_{\text{align-}X} = \frac{1}{|\mathcal{V}_R|}\sum_{i\in\mathcal{V}_R} (1-\cos(\hat{h}_{t,i}, \tilde{h}_{i}))$.

\subsection{Representation entanglement (\reg{}) for retrosynthesis}
\label{sec:reg_method}
Representation entanglement augments the generative state with an auxiliary token $z_t$ and models a joint distribution $p_\theta(G_R,z\mid G_P)$.
We append $z_t$ as an additional global token and model it alongside the graph trajectory.

We consider both discrete and continuous tokens to probe the effect of token bandwidth and semantic granularity.
Discrete tokens (reaction class) provide a low-cost coarse prior available in \USPTOK{}, while continuous tokens (pretrained embeddings) can carry richer reactant semantics.

\paragraph{Discrete token entanglement (reaction class).}
We use the 10-way reaction class label in \USPTOK{} as a discrete token $z_0\in\{1,\dots,10\}$ with a D3PM-style forward transition~\cite{austin2021d3pm}: $q(z_t \mid z_0) = z_0 Q_t$, where $Q_t = (1-\beta_t)I + \beta_t U$.
where $U$ is uniform and $\beta_t$ follows a cosine schedule.
The model predicts logits for the clean token and is optimized with cross-entropy loss $\mathcal{L}_{z}^{\text{disc}}$.

\paragraph{Continuous token entanglement (pretrained embeddings).}
We also entangle continuous vectors from a pretrained encoder (e.g., RMAT CLS or Uni-Mol+ CLS).
Let $z_0\in\mathbb{R}^{d_z}$ be the normalized target embedding.
We apply Gaussian corruption: $z_t = \sqrt{\bar{\alpha}_t}\, z_0 + \sqrt{1-\bar{\alpha}_t}\, \epsilon$, where $\epsilon\sim\mathcal{N}(0,I)$,
and supervise the prediction of $z_0$ with MSE loss $\mathcal{L}_{z}^{\text{cont}}$.

\paragraph{Total training objective.}
$\mathcal{L} = \mathcal{L}_{\RB} + \lambda_{\text{align}}\mathcal{L}_{\text{align}} + \lambda_{z}\mathcal{L}_{z}$.

\subsection{Graph-oriented Representation Guidance (\ours)}
\label{sec:irepa_method}
A plain node-level alignment (Sec.~\ref{sec:repa_method}) aligns each atom token independently through an MLP projector, which ignores bond context.
In molecular graphs, however, chemical semantics is carried by bonded neighborhoods (functional groups, rings), and atom identity is coupled with bond types and valence constraints.
We therefore propose \ours{}, a structure-aware representation alignment that couples node and edge context through a lightweight message-passing projector.

\paragraph{Intermediate features.}
We attach guidance at a Transformer layer $\ell$ (layer 4 for \RB{} by default; Sec.~\ref{sec:layer_results}) and extract
\begin{equation}
X^{(\ell)}_t\in \mathbb{R}^{N\times d_x}, \qquad
E^{(\ell)}_t\in \mathbb{R}^{N\times N\times d_e}.
\end{equation}

\paragraph{Edge augmentation and edge encoder.}
During training, we restrict message passing to edges present in the ground-truth reactant graph.
For each ground-truth bond $(u,v)$, let $b^{\text{GT}}_{uv}\in\{0,1\}^{K_b}$ be the one-hot bond type.
We concatenate it to the intermediate edge features and map to a node-dimensional edge embedding:
\begin{equation}
e^{\text{in}}_{uv}=\text{Concat}\!\left(E^{(\ell)}_{t,uv},\, b^{\text{GT}}_{uv}\right),\quad
e_{uv}=\text{EdgeEnc}\!\left(e^{\text{in}}_{uv}\right),
\end{equation}
where $\text{EdgeEnc}(\cdot)$ is a small MLP that outputs a $d_x$-dimensional edge embedding.

\paragraph{Node--edge coupled projector (EdgeGIN).}
We perform $K$ rounds of message passing along ground-truth neighbors $\mathcal{N}_{\text{GT}}(v)$:
\begin{equation}
\begin{aligned}
h^{(0)}_v &= X^{(\ell)}_{t,v},\\
h^{(k)}_v &= \text{MLP}_k\!\Big((1+\epsilon_k)\,h^{(k-1)}_v + \sum_{u\in\mathcal{N}_{\text{GT}}(v)} \text{ReLU}\!\big(h^{(k-1)}_u + e_{uv}\big)\Big),
\end{aligned}
\end{equation}
for $k=1,\dots,K$ (we use $K=2$), where $\epsilon_k$ is learnable as in GIN~\cite{xu2019gin}.
We denote the resulting structure-aware node features by $\bar{X}_{t,v} := h^{(K)}_v$.

\paragraph{Per-molecule instance normalization on teacher atoms.}
Let $\tilde{h}_i$ denote the teacher atom embedding for atom $i$.
To emphasize within-molecule contrast (relative semantic patterns) rather than absolute embedding scale, we apply instance normalization~\cite{ulyanov2016instance} to teacher atom embeddings within each reactant molecule component:
\begin{equation*}
\tilde{h}'_{i} = \frac{\tilde{h}_{i} - \mu(\tilde{h}_{\mathcal{C}(i)})}{\sigma(\tilde{h}_{\mathcal{C}(i)})+\epsilon},
\end{equation*}
where $\mathcal{C}(i)$ denotes the molecule component containing atom $i$ and $\mu(\cdot),\sigma(\cdot)$ are computed over atoms in that component.
This design highlights the importance of local organization and normalization in the target representation space.

\paragraph{Alignment objective.}
We apply a node-wise head $g_{\phi}$ on top of $\bar{X}_t$ and compute masked cosine alignment on valid reactant atoms:
\begin{equation}
\mathcal{L}_{\text{align-}\ours} =
\frac{1}{|\mathcal{V}_R|}\sum_{i\in\mathcal{V}_R}\left(1-\cos\left(\text{Norm}(g_{\phi}(\bar{X}_{t,i})), \text{Norm}(\tilde{h}'_{i})\right)\right).
\end{equation}

%% file: sections/tts.tex
\section{Inference-time Reranking via Representation Similarity}
\label{sec:rerank_method}

We refine the conventional frequency-based ranking of $n$ samples~\cite{igashov2024retrobridge} by leveraging representation similarity as a lightweight verification score (Fig.~\ref{fig:rerank}). Crucially, this incurs no additional sampling cost or separate training.

\paragraph{Representation similarity score.}
We extract generator-side representations along the denoising trajectory and compute a reference embedding for the candidate $\hat{G}_R$ via the teacher encoder $\mathcal{E}(\cdot)$. We then aggregate per-step cosine similarities over $K$ steps:
\begin{equation}
\label{eq:rerank_sim}
\begin{aligned}
s_t(\hat{G}_R) &= \frac{1}{|\mathcal{V}(\hat{G}_R)|}\sum_{i\in\mathcal{V}(\hat{G}_R)}
\cos\!\Big(\hat{h}_{t,i}(\hat{G}_R), \tilde{h}_{i}(\hat{G}_R)\Big), \\
s(\hat{G}_R) &= \frac{1}{K}\sum_{t=T-K+1}^{T} \frac{s_t(\hat{G}_R) + 1}{2},
\end{aligned}
\end{equation}
where $\cos(\cdot,\cdot)$ is cosine similarity; $T$ is the total number of denoising steps; and $\frac{s_t+1}{2}$ maps cosine similarity to $[0,1]$ for numerical stability.
$\mathcal{V}(\hat{G}_R)$ denotes the set of representation tokens used for similarity.
$\tilde{h}_{i}(\hat{G}_R)$ denotes the teacher-side embedding of node $i$ on $\hat{G}_R$, and $\hat{h}_{t,i}(\hat{G}_R)$ denotes the generator-side representation (at step $t$) of node $i$ on $\hat{G}_R$(Appendix~\ref{sec:appendix_rerank}).

\paragraph{Frequency score.}
Let $c(\hat{G}_R)\in\{1,\dots,n\}$ denote the number of times a distinct candidate $\hat{G}_R$ appears among the $n$ sampled trajectories.
We define the frequency score as $f(\hat{G}_R) = \frac{c(\hat{G}_R)}{n}\in[0,1]$,
which is the default confidence signal used by frequency ranking~\cite{igashov2024retrobridge}.

\paragraph{Reranking score via fusion.}
We combine $f(\hat{G}_R)$ and $s(\hat{G}_R)$ with a simple fusion rule to obtain a final reranking score $\text{Score}(\hat{G}_R)$ (implementation details in Appendix~\ref{sec:impl}).
Intuitively, $f(\hat{G}_R)$ captures model confidence under stochastic sampling, while $s(\hat{G}_R)$ acts as a semantic consistency score that helps break ties among high-frequency candidates.

%% file: sections/results.tex
\section{Empirical Results and Design Insights}
\label{sec:results}

We evaluate representation guidance on \USPTOK{} and an out-of-distribution (OOD) cluster split~\cite{qiao2025rarb},
using \RB{} as the primary stochastic-process backbone.
We start from the most broadly applicable instantiation,
identify accuracy--diversity trade-off in one-to-many retrosynthesis, and then make
a sequence of explicit \emph{decision points}.
These decisions establish a strong and stable baseline recipe. Building on this foundation, we \emph{carefully tailor} representation guidance
to molecular graphs by explicitly accounting for node--bond coupling and structure-aware
correspondence under categorical corruption, culminating in our final method,
\ours{} (Graph-oriented Representation Guidance).

\subsection{Graph-level alignment and teacher choice}
\label{sec:teacher_results}

Graph-level alignment on the denoiser global token $y$ is the most broadly applicable entry point.
Table~\ref{tab:topk_main} reports a consolidated comparison across teacher.

\begin{table*}[t]
\centering
\small
\caption{Top-$k$ exact-match accuracy (\%) on the \USPTOK{} test split.}
\label{tab:topk_main}
\begin{tabular}{l l l cccc c}
\toprule
Model & Guidance granularity & Teacher / token & Top-1 & Top-3 & Top-5 & Top-10 & Diversity \\
\midrule
\RB{} (Base) & -- & -- & 50.8 & 71.1 & 76.0 & 80.3 & 12.9 \\
\midrule
RB-Align$_y$ & graph $y$ & Uni-RXN (prod+reac) & 56.8 & 72.7 & 76.0 & 78.4 & 10.9 \\
RB-Align$_y$ & graph $y$ & Uni-RXN (reac-only) & 57.6 & 73.1 & 76.6 & 78.8 & 11.8 \\
RB-Align$_y$ & graph $y$ & ECFP/Morgan & 56.6 & 71.3 & 75.2 & 78.6 & 11.1 \\
RB-Align$_y$ & graph $y$ & RMAT & 57.0 & 72.5 & 74.8 & 77.4 & 9.3 \\
\midrule
RB-Align$_X$ & node $X$ & Uni-RXN node & 58.0 & 76.6 & 80.2 & 85.3 & 13.7 \\
\midrule
\ours{} & node--edge & Uni-RXN node & \textbf{58.6} & \textbf{77.2} & \textbf{83.4} & \textbf{87.1} & \textbf{15.5} \\
\bottomrule
\end{tabular}
\end{table*}

\noindent\textbf{Takeaway.}
Across teacher types, Align$_y$ substantially improves top-1 accuracy over the \RB{} baseline
(Table~\ref{tab:topk_main}), indicating that representation transfer is effective even without a
CLIP-level universally dominant molecular encoder.
At the same time, Uni-RXN is the strongest among tested teachers, suggesting that reaction-aware representations better match retrosynthesis semantics than generic
molecule encoders or fingerprints.

\noindent\textbf{But: graph-level alignment exhibits an accuracy--diversity trade-off.}
While Align$_y$ improves small-$k$ precision, it reduces diversity (and can hurt larger-$k$ coverage),
foreshadowing over-constraint in one-to-many reaction spaces.

\noindent\textbf{Decision.}
We use Uni-RXN which is a \emph{reaction-aware} pretrained encoder~\cite{qiang2023unirxn} as the default teacher.

\subsection{From global constraints to local guidance in one-to-many retrosynthesis}
\label{sec:results_main}

With the teacher fixed (Uni-RXN), we now decide \emph{where} to apply guidance (endpoint),
\emph{at what granularity} (graph vs.\ node), and \emph{how deep} to inject it (layer placement).

\noindent\textbf{Decision point I: endpoint choice for graph-level alignment.}
Comparing Uni-RXN dual-endpoint alignment vs.\ reactant-only alignment
(Table~\ref{tab:topk_main}), we observe that aligning the generated endpoint only
(reactant-only) yields a better accuracy--diversity trade-off.
Intuitively, the product endpoint is already observed as conditioning; additionally constraining it can unnecessarily narrow the conditional distribution.

\noindent\textbf{Decision.}
We adopt \textbf{reactant-only} guidance.

\subsubsection{Multi-reaction-center analysis}
\label{sec:rc_results}

We further stratify performance by reaction-center (RC) complexity.
Detailed numbers are provided in Appendix Table~\ref{tab:rc}.

\noindent\textbf{Design insight: coarse global constraints can collapse large-$k$ performance on harder (RC$>$1) reactions.}
Graph-level alignment shows strong top-1 gains but degrades larger-$k$ accuracy on RC$>$1 reactions,
consistent with the notion that global constraints pick a single ``high-confidence'' mode and miss
alternative valid decompositions.

\noindent\textbf{Decision point II: move from graph-level to node-level alignment.}
To attempt more fine-grained guidance, we move to node-level
supervision using Uni-RXN atom embeddings.
As shown by RB-Align$_X$ in Table~\ref{tab:topk_main}, node-level alignment not only improves top-$k$
across the board but also \emph{recovers and exceeds} baseline diversity.
Moreover, it substantially mitigates the large-$k$ degradation on harder multi-reaction-center
(RC$>$1) reactions that we observe with coarse graph-level alignment
(Appendix Table~\ref{tab:rc}).
This suggests that distributing guidance across atoms provides more informative, localized
constraints without globally collapsing the reactant distribution.

\noindent\textbf{Decision.}
We adopt \textbf{node-level} reactant-only alignment.

\subsubsection{Layer placement for node-level alignment}
\label{sec:layer_results}

Having decided to use node-level alignment, the remaining practical question is \emph{where} to
inject it in the denoiser.
We report a layer-placement ablation in Appendix Table~\ref{tab:layer}.

\noindent\textbf{Design insight.}
Aligning deeper layers consistently improves node-level results (Appendix Table~\ref{tab:layer}), suggesting
that deeper denoiser features encode more task-relevant chemical semantics.

\noindent\textbf{Decision.}
We use \textbf{layer 4} for node-level alignment.

\subsection{Robustness and transfer}
\label{sec:robustness_alignment}

At this point we have a concrete, fixed alignment recipe.
We now evaluate whether this choice improves (i) plausibility under forward-model checks,
(ii) OOD generalization, and (iii) transfer beyond Markov bridges.

\subsubsection{Round-trip consistency and diversity}
\label{sec:roundtrip}

Table~\ref{tab:roundtrip} reports round-trip coverage and accuracy using the Molecular Transformer
forward model~\cite{schwaller2019moleculartransformer,schwaller2020pathways}.

\begin{table}[t]
\centering
\small
\caption{Round-trip coverage and accuracy (\%) on the standard test split of \USPTOK{}.}
\label{tab:roundtrip}
\begin{tabular}{l ccc ccc}
\toprule
& \multicolumn{3}{c}{Coverage} & \multicolumn{3}{c}{Accuracy} \\
Model & k=1 & k=3 & k=5 & k=1 & k=3 & k=5 \\
\midrule
\RB{} (Base) & 84.2 & 94.3 & 95.9 & 84.2 & 71.7 & 66.3 \\
RB-Align$_y$ (Uni-RXN) & 90.7 & 97.2 & 97.8 & 90.7 & 76.4 & 70.9 \\
RB-Align$_X$ (Uni-RXN node) & 89.3 & 97.8 & 99.0 & 89.3 & 78.7 & 72.7 \\
\ours{} (node--edge) & 92.1 & 98.4 & 99.4 & 92.1 & 79.9 & 72.7 \\
\bottomrule
\end{tabular}
\end{table}

\noindent\textbf{Takeaway.}
RB-Align$_X$ improves round-trip coverage compared to the \RB{} baseline, especially at larger $k$
(Table~\ref{tab:roundtrip}), indicating that the guided generator produces candidates that are not
only closer to the ground truth but also more often forward-consistent.

\subsubsection{Out-of-distribution generalization on \textsc{USPTO-50k-cluster}}
\label{sec:ood_results}

We evaluate OOD generalization on \textsc{USPTO-50k-cluster}~\cite{qiao2025rarb}.

\begin{table}[t]
\centering
\small
\caption{Top-k (exact match) accuracy on the \textsc{USPTO-50k-cluster} dataset’s test split.}
\label{tab:ood_cluster}
\begin{tabular}{l cccc}
\toprule
Model & Top-1 & Top-3 & Top-5 & Top-10 \\
\midrule
\RB{} (Base) & 42.9 & 66.3 & 73.2 & 77.9 \\
RB-Align$_X$ (node) & 54.7 & 69.9 & 76.6 & 79.3 \\
\ours{} (node--edge) & 55.3 & 70.7 & 76.8 & 80.3 \\
\bottomrule
\end{tabular}
\end{table}

\noindent\textbf{Takeaway: node-level alignment provides a strong, transferable semantic anchor.}
RB-Align$_X$ substantially improves OOD top-$k$ accuracy across all $k$
(Table~\ref{tab:ood_cluster}), with a particularly large gain on top-1 (42.9$\rightarrow$54.7).
This supports that representation alignment helps distill intrinsic chemical
semantics rather than overfitting to
dataset-specific shortcuts.

\subsubsection{Transfer to a flow model: \synbridge{}}
\label{sec:results_synbridge}

To test whether the benefits of representation alignment extend beyond Markov bridges, we inject an
analogous alignment mechanism into \synbridge{}~\cite{lin2025synbridge}.
Results are summarized in Appendix Table~\ref{tab:synbridge} under the same evaluation pipeline
(Appendix~\ref{sec:metrics}).

\noindent\textbf{Takeaway.}
On \synbridge{}, alignment improves exact-match top-$k$ yet reduces round-trip coverage
(Appendix Table~\ref{tab:synbridge}), suggesting that the interaction between guidance and the underlying
stochastic process and architecture can shift the precision--plausibility balance.
This motivates comparing guidance schemes beyond alignment.

\subsection{Guidance scheme under categorical corruption: entanglement vs.\ alignment}
\label{sec:reg_results}

We next test representation entanglement (\reg{}) by jointly modeling an auxiliary token
$z$ appended to the global representation.
We consider both discrete tokens (reaction class) and continuous tokens (pretrained CLS embeddings)
to probe how token bandwidth and modality affect one-to-many behavior.

\begin{table*}[t]
\centering
\small
\caption{Representation entanglement via joint modeling of auxiliary token and graph. Exact-match accuracy (\%) on \USPTOK{}.}
\label{tab:reg}
\begin{tabular}{l l c cccc c}
\toprule
Model & Token type & Dim & Top-1 & Top-3 & Top-5 & Top-10 & Diversity \\
\midrule
\RB{} (Base) & -- & -- & 50.8 & 71.1 & 76.0 & 80.3 & 12.9 \\
\midrule
RB-REG (reaction type) & discrete one-hot & 10 & 60.0 & 74.1 & 78.0 & 80.4 & 11.7 \\
RB-REG (RMAT CLS) & continuous CLS & 64 & 58.8 & 77.0 & 80.8 & 83.8 & 13.2 \\
RB-REG (Uni-Mol+ CLS) & continuous CLS & 64 & 59.2 & 77.8 & 81.0 & 83.2 & 10.3 \\
RB-REG (Uni-RXN CLS) & continuous CLS & 64 & 58.6 & 74.6 & 78.0 & 80.2 & 9.8 \\
RB-REG (Uni-RXN CLS) & continuous CLS & 16 & 58.4 & 73.5 & 76.6 & 79.0 & 10.9 \\
\bottomrule
\end{tabular}
\end{table*}

\noindent\textbf{Takeaway: entanglement can improve small-$k$, but exhibits sharper token-dependent trade-offs.}
Entanglement variants improve top-1 in several settings (Table~\ref{tab:reg}), and the discrete
reaction-type token achieves the highest top-1 among tested configurations.
However, entanglement often reduces diversity, indicating that coupling an auxiliary token into the
generative state can impose strong constraints in one-to-many reaction spaces.

\noindent\textbf{Design insight: token modality and source matter.}
Pretrained continuous tokens outperform learned/random tokens (Appendix Table~\ref{tab:reg_cls_source}),
suggesting that entanglement benefits from meaningful external semantics; nevertheless, selecting a
token that preserves diversity remains non-trivial.

\noindent\textbf{Decision.}
Given its more stable accuracy--diversity behavior, we use
\textbf{alignment} as the default guidance scheme for our final design.

\subsection{Graph-oriented Representation Guidance}
\label{sec:grg_results}

So far, we have established a stable alignment recipe that improves both in-distribution accuracy and OOD generalization.
The remaining question is \emph{graph-specific}: can we transfer semantics more reliably by
explicitly accounting for node--edge coupling, rather than aligning atom tokens in isolation?

\noindent\textbf{\ours{}: structure-aware alignment via node--edge coupling.}
\ours{} (Sec.~\ref{sec:irepa_method}) replaces the plain per-node MLP projector with a lightweight
message-passing projector that aggregates bonded neighborhoods (EdgeGIN) and applies per-molecule
instance normalization on teacher atom embeddings.

\noindent\textbf{Main result: \ours{} improves both accuracy \emph{and} diversity.}
As shown in Table~\ref{tab:topk_main}, \ours{} consistently improves all top-$k$ metrics over
RB-Align$_X$ while further increasing diversity (15.5 vs.\ 13.7), indicating that incorporating graph
structure into representation transfer strengthens guidance without collapsing one-to-many modes.

\noindent\textbf{Robustness:}
\ours{} achieves the highest round-trip coverage (Table~\ref{tab:roundtrip}) and is consistently
strongest on the OOD cluster split across all $k$ (Table~\ref{tab:ood_cluster}).
These improvements are particularly meaningful because they persist under distribution shift, which
suggests that \ours{} better learns crucial and intrinsic chemical semantics.

\subsection{Inference-time reranking via representation similarity}
\label{sec:rerank_results}

We evaluate inference-time reranking that fuses the default frequency score with the
representation-similarity score described in Sec.~\ref{sec:rerank_method}.
We summarize reranking as $\Delta$Top-$k$ relative to the default frequency ranking.
Detailed numbers are provided in Appendix Table~\ref{tab:rerank}

\noindent\textbf{Takeaway: guided representations enable lightweight verification signals.}
Representation similarity most consistently improves the head of the ranked list (often top-1) (Appendix Table~\ref{tab:rerank}).
For \ours{}, reranking improves top-1/top-3/top-10 without any decrease, suggesting that the guided
representation space provides a useful semantic consistency signal.

\subsection{Compute and efficiency}
\label{sec:efficiency}

\noindent\textbf{Faster convergence in epochs and wall-clock time.}
\ours{} reduces the number of epochs by 35\% and the wall-clock time by 30\% to reach comparable
validation performance. The full validation curves are provided in Appendix~\ref{sec:appendix_efficiency}
+(Fig.~\ref{fig:eff_epochs}--\ref{fig:eff_wallclock}).

\noindent\textbf{Training overhead.}
Alignment introduces a modest training overhead (Appendix~\ref{sec:impl}).
Our structure-aware projector in \ours{} incurs a slightly higher but still low overhead.
Entanglement increases training cost further due to the additional token pathway.

\noindent\textbf{Sampling and reranking overhead.}
We can do without additional sampling budget when using alignment.
Reranking logic and sampling cls tokens are negligible; practical overhead mainly comes from teacher-encoder throughput when
computing candidate-side references for similarity.

%% file: sections/limitations.tex
\section{Limitations and Ethical Considerations}
\label{sec:limitations}

\noindent\textbf{Limitations.}
Our improvements depend on the teacher encoder; teacher failure modes may be inherited by alignment and similarity-based reranking. We focus on \USPTOK{} (plus one OOD split and one architecture transfer), and broader validation on larger and more diverse reaction corpora remains future work. 

\noindent\textbf{Limitations: \reg{} design space.}
We only probe a small set of auxiliary tokens and corruption settings for representation entanglement; richer tokenizations and schedules may yield better performance.

\noindent\textbf{Ethical considerations.}
We utilize the public \USPTOK{} dataset, involving no human subjects or sensitive data. regarding dual-use risks, while improved retrosynthesis lowers barriers for synthesizing regulated compounds, we emphasize the necessity of safeguards in any model deployment.

%% file: sections/conclusion.tex
\section{Conclusion}

We presented \ours{}, a graph-oriented representation guidance framework for template-free retrosynthesis with stochastic-process molecular-graph generators.
Across a unified empirical study spanning teacher representations, guidance endpoints and granularity, injection depth, and graph-specific structure-aware designs under categorical corruption, we distill the practical choices that make representation guidance reliable for one-to-many retrosynthesis.
Building on these insights, \ours{} couples node and bond context via a lightweight message-passing projector for stable, structure-aware alignment, achieving 58.6\% top-1 / 87.1\% top-10 accuracy with higher diversity on \USPTOK{}, consistent improvements under distribution shift, and faster convergence in both epochs and wall-clock time.
Moreover, the guided representation space provides a simple post-sampling reranking signal via representation similarity, improving the head of the ranked list without training an additional verifier.
Overall, our results indicate that explicit representation guidance is a practical path to distill more intrinsic chemical abstractions into discrete graph generators for synthesis planning.

%% file: sections/appendix.tex
\section{Additional Details}

\subsection{Evaluation metrics and protocol}
\label{sec:metrics}

\paragraph{Top-$k$ exact match.}
Top-$k$ exact-match accuracy is the fraction of products for which the ground-truth reactants appear within the top-$k$ \emph{distinct} ranked candidates (higher is better).
At test time we sample $n=\nsamples$ candidates per product, remove duplicates, and rank candidates by frequency unless otherwise noted.

\paragraph{Round-trip correctness.}
We use the forward Molecular Transformer~\cite{schwaller2019moleculartransformer,schwaller2020pathways}.
Let $F(\cdot)$ denote the forward model.
A predicted reactant set is forward-consistent if $F(G_R)=G_P$.
A prediction is treated as correct if it either exactly matches the ground truth or is forward-consistent.

\paragraph{Round-trip coverage and accuracy.}
Round-trip coverage@k measures the fraction of products for which at least one of the top-$k$ distinct ranked predictions is correct (higher is better).
Round-trip accuracy@k measures the fraction of correct predictions among the selected top-$k$ predictions (higher is better).

\paragraph{Diversity.}
We report diversity as the average number of distinct reactant candidates among the $n=\nsamples$ sampled predictions per product (higher is better).

\subsection{OOD split: \textsc{USPTO-50k-cluster}}
\label{sec:appendix_cluster}
We evaluate out-of-distribution generalization on the \textsc{USPTO-50k-cluster} split~\cite{qiao2025rarb}.
This split is designed to reduce overlap between training and test chemistry.
We use the provided train/valid/test partitions and report test results in Table~\ref{tab:ood_cluster}.

\subsection{Implementation details}
\label{sec:impl}

\paragraph{Backbone implementations.}
\noindent\textbf{\RB{} (Markov bridge).}
Our primary backbone is \RB{}~\cite{igashov2024retrobridge}, which formulates retrosynthesis as sampling from a discrete-time Markov bridge process $\{G_t\}_{t=0}^T$ with $T=\Tsteps$ steps, where $G_T$ is the fixed product endpoint and $G_0$ is the reactant target.
A Graph Transformer denoiser $D_\theta(G_t,t,G_P)$ learns to reverse the forward corruption / transport.
At each layer $\ell$, the denoiser exposes intermediate node states $X^{(\ell)}_t$, edge states $E^{(\ell)}_t$, and a global token $y^{(\ell)}_t$; our alignment/entanglement heads attach to these intermediate states.

\paragraph{Padded graphs, dummy nodes, and node order.}
RetroBridge uses a fixed-size dense graph representation with \emph{dummy} nodes: because some atoms present in the reactants may be absent in the corresponding product, the official setup always adds 10 dummy nodes to each initial product graph~\cite{igashov2024retrobridge}.
The implementation also randomly permutes graph nodes after converting SMILES to graphs to avoid ordering shortcuts.
We keep (i) a binary node-validity indicator $m$ to exclude padded/dummy slots when computing node-level alignment losses, and (ii) a stored index vector that maps the permuted student node order back to the teacher's canonical atom order.
Along the bridge trajectory, dummy slots can ``birth'' into atoms (or ``die'' back into dummy); by applying atom-level alignment only on ground-truth reactant atoms, we avoid spurious supervision from dummy slots.

\noindent\textbf{\synbridge{} (discrete flow).}
For architectural generalization, we additionally inject alignment into \synbridge{}~\cite{lin2025synbridge}.
We follow the same teacher preprocessing and endpoint choice as in our graph-level alignment setting.

\paragraph{Teacher encoders.}
We evaluate Uni-RXN~\cite{qiang2023unirxn}, ECFP/Morgan~\cite{rogers2010ecfp}, RMAT~\cite{maziarka2021rmat}, and Uni-Mol+~\cite{zhou2023unimol}.
Teacher embeddings used for supervision are precomputed and cached for efficiency.

\paragraph{Embedding preprocessing.}
Graph-level teacher embeddings are PCA-whitened to a fixed dimension and then $L_2$-normalized.
For Uni-RXN atom embeddings used in node-level alignment, we apply PCA whitening and normalization at the atom level (dimensions reported in the main paper where relevant).

\paragraph{Guidance heads.}
\noindent\textbf{Graph-level alignment heads.}
We use a small MLP projection head on top of the global token $y$ and optimize cosine alignment.

\noindent\textbf{Node-level alignment heads.}
We use a node-wise MLP head on top of intermediate node features and optimize masked cosine alignment.

\paragraph{Training.}
We train with AdamW (amsgrad=True), learning rate $2\times 10^{-4}$, weight decay $10^{-12}$, and no explicit learning-rate schedule.
We use batch size 128, seed 42, and train for 1000 epochs.
For \RB{}, we use a cosine noise schedule with $T=\Tsteps$ steps.

\paragraph{Inference and reranking.}
At test time, we sample $n=\nsamples$ reactant candidates per product.
Default ranking is sample frequency among $n$ draws.
Inference-time reranking requires encoding generated candidates with a pretrained encoder to obtain similarity references (Sec.~\ref{sec:rerank_method}); this is the main source of reranking overhead.

\begin{table}[t]
\centering
\small
\caption{Compute overhead summary \emph{relative to} \RB{} (Base) in our implementation. ``Training cost'' is normalized by average training time per epoch; ``Sampling cost'' is normalized by end-to-end time for generating the default sampling budget (reranking excluded).}
\label{tab:overhead}
\begin{tabular}{l c c}
\toprule
Model & Training cost (rel.) & Sampling cost (rel.) \\
\midrule
\RB{} (Base) & 1.00$\times$ & 1.00$\times$ \\
RB-Align$_X$ (alignment) & 1.08$\times$ & 1.00$\times$ \\
\ours{} (node--edge) & 1.10$\times$ & 1.00$\times$ \\
RB-REG (RMAT CLS) & 1.13$\times$ & 1.01$\times$ \\
\bottomrule
\end{tabular}
\end{table}
\subsection{Additional ablations and analyses}
\label{sec:appendix_additional_ablations}
This section collects supplementary ablations and analyses referenced in the main text.

\begin{table}[t]
\centering
\small
\caption{Percentage-point change in top-$k$ exact-match accuracy relative to the \RB{} baseline, stratified by number of reaction centers (RC).}
\label{tab:rc}
\begin{tabular}{l cccc}
\toprule
Model / Split & Top-1 & Top-3 & Top-5 & Top-10 \\
\midrule
RB-Align$_y$ (Uni-RXN) RC=1 & 5.9 & 2.3 & 0.9 & -0.9 \\
RB-Align$_y$ (Uni-RXN) RC$>$1 & 7.1 & -11.3 & -16.4 & -20.4 \\
\midrule
RB-Align$_X$ (Uni-RXN node) RC=1 & 7.4 & 6.0 & 4.5 & 5.6 \\
RB-Align$_X$ (Uni-RXN node) RC$>$1 & 3.4 & -3.9 & -1.6 & -5.6 \\
\bottomrule
\end{tabular}
\end{table}

\begin{table}[t]
\centering
\small
\caption{Layer placement ablation for node-level alignment.}
\label{tab:layer}
\begin{tabular}{l cccc}
\toprule
Alignment layer & Top-1 & Top-3 & Top-5 & Top-10 \\
\midrule
2nd layer & 56.2 & 73.7 & 78.0 & 81.4 \\
3rd layer & 57.2 & 75.4 & 80.2 & 83.2 \\
4th layer & 58.0 & 76.6 & 80.2 & 85.3 \\
\bottomrule
\end{tabular}
\end{table}

\begin{table}[t]
\centering
\small
\caption{\synbridge{} generalization. Alignment improves exact-match top-$k$ in this setting.}
\label{tab:synbridge}
\begin{tabular}{l cccc | cccc}
\toprule
 & \multicolumn{4}{c|}{Exact match} & \multicolumn{4}{c}{Round-trip coverage} \\
Model & 1 & 3 & 5 & 10 & 1 & 3 & 5 & 10 \\
\midrule
\synbridge{} (Base) & 78.7 & 82.3 & 83.8 & 84.7 & 91.8 & 94.1 & 94.8 & 95.2 \\
\synbridge{} (Align) & 81.4 & 83.6 & 84.7 & 85.4 & 83.4 & 85.4 & 86.4 & 87.0 \\
\bottomrule
\end{tabular}
\end{table}

\begin{table}[t]
\centering
\small
\caption{Representation entanglement ablation on source.}
\label{tab:reg_cls_source}
\begin{tabular}{l cccc c}
\toprule
CLS token source & Top-1 & Top-3 & Top-5 & Top-10 & Diversity \\
\midrule
Random vector (learned) & 57.4 & 73.3 & 76.8 & 78.6 & 10.8 \\
Pooled node embedding & 58.0 & 74.3 & 77.8 & 79.4 & 9.0 \\
Pretrained CLS (RMAT CLS) & 58.8 & 77.0 & 80.8 & 83.8 & 13.2 \\
\bottomrule
\end{tabular}
\end{table}

\subsection{Efficiency curves}
\label{sec:appendix_efficiency}
We provide the validation exact-match curves referenced in Sec.~\ref{sec:efficiency}.

\begin{figure}[t]
\centering
\includegraphics[width=0.92\linewidth]{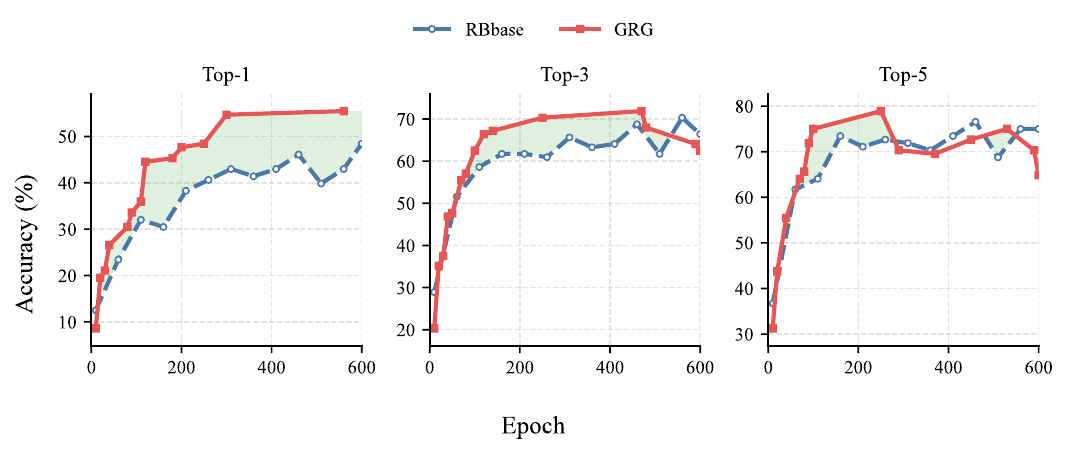}
\caption{Validation exact-match vs.\ epochs for \RB{} (Base) and \ours{}.}
\label{fig:eff_epochs}
\end{figure}

\begin{figure}[t]
\centering
\includegraphics[width=0.92\linewidth]{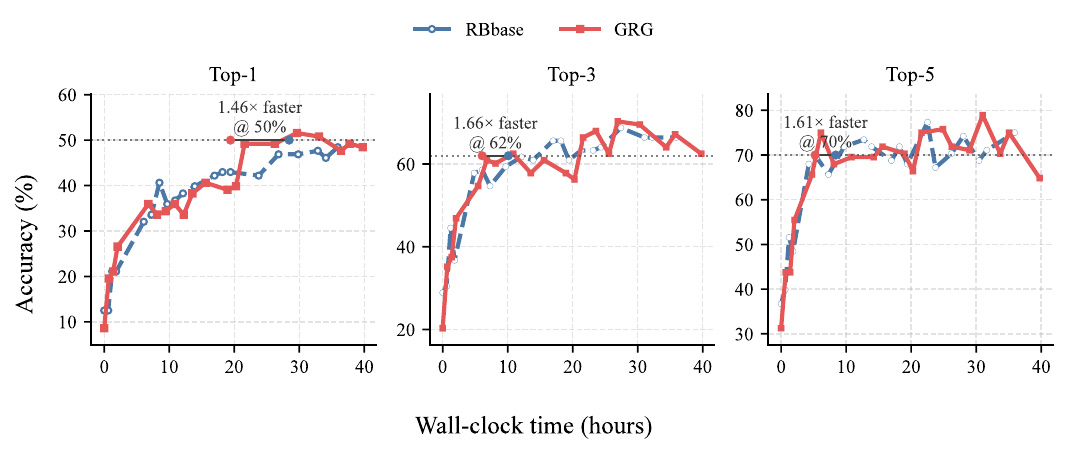}
\caption{Validation exact-match vs.\ wall clock for \RB{} (Base) and \ours{}.}
\label{fig:eff_wallclock}
\end{figure}

\subsection{Alignment losses and masking details}
For node-level alignment, dummy atoms introduced by padding must be excluded from the alignment objective.
We use a binary mask $m\in\{0,1\}^N$ indicating valid atoms and compute a masked mean of per-node cosine-alignment losses (Sec.~\ref{sec:repa_method}).
In all node-level experiments we align only on the reactant side, following the empirical finding that reactant-only alignment yields a better accuracy--diversity trade-off.

\subsection{Node ordering bookkeeping for atom-level alignment}
\label{sec:appendix_order}
Our atom-level teacher (Uni-RXN) produces per-atom embeddings in RDKit atom-index order.
In contrast, RetroBridge constructs the reactant-graph node order by first sorting atoms by atom-map numbers (to define a canonical ``mapped'' order) and then applying a random permutation; the padded graph may also include dummy nodes.
To align these two orderings during training, the dataset preprocessing stores an index vector $\texttt{reactant\_order}\in\{0,\dots,N_{\text{atoms}}-1\}^N$ such that the model's node position $j$ corresponds to RDKit atom index $\texttt{reactant\_order}[j]$ (dummy nodes are masked out by $m_j=0$).
We therefore reorder teacher targets to model order as
$H^{\text{reord}}_j = H^{\text{teacher}}_{\texttt{reactant\_order}[j]}$ and apply masked cosine alignment only on valid nodes.
No atom mapping or reordering is needed at inference, since alignment is used only as training-time supervision.

\subsection{Additional implementation notes for \ours{} (structure-aware projector)}
\label{sec:appendix_irepa}
Our structure-aware alignment (Sec.~\ref{sec:irepa_method}) uses a lightweight EdgeGIN-style message-passing projector:
\begin{itemize}[leftmargin=*]
  \item \textbf{Inputs:} intermediate node features $X^{(\ell)}_t$, intermediate edge features $E^{(\ell)}_t$, and a one-hot ground-truth bond type $\text{Bond}^{\text{GT}}$.
  \item \textbf{Training-only privileged filtering:} message passing is restricted to ground-truth reactant edges (computed from the clean reactant graph).
  \item \textbf{Inference:} privileged information is used only inside the auxiliary alignment projector during training; the generative pathway and inference do not depend on it.
  \item \textbf{Normalization:} we apply per-molecule instance normalization on teacher embeddings to emphasize within-molecule token-level contrast and improve alignment stability.
\end{itemize}

\subsection{Inference-time reranking: similarity computation and variants}
\label{sec:appendix_rerank}

\paragraph{Score fusion.}
We use a simple linear fusion of the frequency score $f(\hat{G}_R)\in[0,1]$ and the
representation-similarity score $s(\hat{G}_R)\in[0,1]$:
\begin{equation*}
\text{Score}(\hat{G}_R)=0.85\,f(\hat{G}_R)+0.15\,s(\hat{G}_R).
\end{equation*}
Candidates are ranked by $\text{Score}(\hat{G}_R)$ in descending order (ties are rare; when they
occur we break ties by higher $f$).
\emph{Example:} if a distinct candidate appears $c=20$ times among $n=100$ samples, then
$f=0.20$. If its similarity score is $s=0.74$, then
$\text{Score}=0.85\times 0.20+0.15\times 0.74=0.281$.

\begin{table}[t]
\centering
\small
\caption{Effect of inference-time reranking via representation similarity on \USPTOK{}.}
\label{tab:rerank}
\begin{tabular}{l cccc}
\toprule
Model & $\Delta$Top-1 & $\Delta$Top-3 & $\Delta$Top-5 & $\Delta$Top-10 \\
\midrule
RB-Align$_y$ (Uni-RXN) & +0.8 & +0.4 & +0.2 & +0.2 \\
RB-Align$_y$ (ECFP) & -0.6 & +0.2 & +0.2 & +0.0 \\
RB-Align$_y$ (RMAT) & +0.8 & -0.4 & +0.2 & +0.0 \\
RB-Align$_X$ (Uni-RXN node) & +1.4 & +0.0 & +0.2 & -0.3 \\
\ours{} & +0.6 & +1.0 & +0.0 & +0.4 \\
RB-REG (RMAT CLS) & +1.4 & +0.2 & +0.0 & -0.4 \\
RB-REG (Uni-RXN CLS) & +0.6 & +0.2 & -0.6 & +0.0 \\
\bottomrule
\end{tabular}
\end{table}

\paragraph{Similarity computation.}
Reranking fuses the default frequency score with a representation-similarity score computed from the guided representations (Sec.~\ref{sec:rerank_method}).
For each sampled trajectory we extract generator-side guidance-head representations along the denoising trajectory and compute teacher-side representations using the same teacher encoder and the same normalization used in training (PCA whitening + $L_2$ normalization; Appendix~\ref{sec:impl}).
We then compute the per-step similarity $s_t(\hat{G}_R)$ in Eq.~\eqref{eq:rerank_sim} by averaging cosine similarity over the token set $\mathcal{V}(\hat{G}_R)$:
\begin{itemize}[leftmargin=*]
  \item \textbf{Graph-level guided models.} $\mathcal{V}(\hat{G}_R)$ is a singleton global token; $\hat{h}_{t,i}$ is the projected global token (e.g., $y_t^{(\ell)}$), and $\tilde{h}_i$ is the teacher set embedding computed with the same pooling as in Eq.~\eqref{eq:set_embed}.
  \item \textbf{Node-level guided models.} $\mathcal{V}(\hat{G}_R)$ indexes the valid atoms in the decoded reactant graph; $\hat{h}_{t,i}$ is the projected node embedding at atom $i$, and $\tilde{h}_i$ is the teacher atom embedding aligned to the same node index.
        When node-level similarity is used, we align teacher atom embeddings to the graph node order by applying the same graph construction and node-order bookkeeping routine used during training (Appendix~\ref{sec:appendix_order}).
\end{itemize}

\paragraph{Duplicates among sampled candidates.}
If a distinct candidate $\hat{G}_R$ appears multiple times among the $n$ sampled trajectories, we compute its similarity score per occurrence and average the resulting $s(\hat{G}_R)$ before fusing with the frequency score.

\paragraph{Aggregation window.}
The results in Table~\ref{tab:rerank} use a similarity signal aggregated over the last $K=10$ denoising steps.
Additional variants are possible:
\begin{itemize}[leftmargin=*]
  \item Using only the last-step similarity.
  \item Averaging similarity across a different window size (e.g., last $K$ steps with $K\neq 10$).
  \item Using both a trajectory-average similarity and a late-step similarity in a two-term score.
  \item Learning a small calibration map for similarity based on validation data, while keeping inference-time reranking label-leakage-safe.
\end{itemize}
We leave systematic exploration of these variants to future work.